\documentclass{llncs}
\usepackage{llncsdoc}

\usepackage{tabularx}

\usepackage{times}
\usepackage{graphicx}
\usepackage{array}
\usepackage{amsmath}
\usepackage{amssymb}
\usepackage{algorithm}
\usepackage{algorithmic}
\usepackage{url}
\usepackage{multirow}
\usepackage{subfigure}
\usepackage{sidecap}

\usepackage[export]{adjustbox}

\newcolumntype{V}{>{$\vcenter\bgroup\hbox\bgroup}c<{\egroup\egroup$}}
\def\Hline{\noalign{\hrule height 4\arrayrulewidth}}

\graphicspath{{./imgs/}}

\begin{document}

\title{A Unified Framework for Thermal Face Recognition}
\author{Reza Shoja Ghiass$^\dag$, Ognjen Arandjelovi{\'c}$^\ddag$, Hakim Bendada$^\dag$, and Xavier Maldague$^\dag$}
\institute{$^\dag$ Laval University, Canada \hspace{50pt}$^\ddag$ Deakin University, Australia\vspace{-20pt}}

\maketitle

\begin{abstract}
The reduction of the cost of infrared (IR) cameras in recent years has made IR imaging a highly viable modality for face recognition in practice. A particularly attractive advantage of IR-based over conventional, visible spectrum-based face recognition stems from its invariance to visible illumination. In this paper we argue that the main limitation of previous work on face recognition using IR lies in its \textit{ad hoc} approach to treating different nuisance factors which affect appearance, prohibiting a unified approach that is capable of handling concurrent changes in multiple (or indeed all) major extrinsic sources of variability, which is needed in practice. We describe the first approach that attempts to achieve this -- the framework we propose achieves outstanding recognition performance in the presence of variable (i) pose, (ii) facial expression, (iii) physiological state, (iv) partial occlusion due to eye-wear, and (v) quasi-occlusion due to facial hair growth.
\end{abstract}

\section{Introduction}
\label{s:intro}
Following the overly optimistic expectations of early research, after several decades of intense research on the one hand and relatively disappointing practical results on the other, face recognition technology has been finally started to enjoy some success in the consumer market. An example can be found within the online photo sharing platform in Google Plus which automatically recognizes individuals in photographs based on previous labelling by the user. It is revealing to observe that the recent success of face recognition has been in the realm of data and image retrieval \cite{Shan2010}, rather than security, contrasting the most often stated source of practical motivation driving past research \cite{ChelWilsSiro1995}. This partial paradigm shift is only a recent phenomenon. In hindsight, that success would first be achieved in the domain of retrieval should not come as a surprise considering the relatively low importance of type-II errors in this context: the user is typically interested in only a few of the retrieved matches and the consequences of the presence of false positives is benign, more often being mere inconvenience.


Nevertheless, the appeal of face recognition as a biometric means that the interest in its security applications is not waning. Face recognition can be performed from a distance and without the knowledge of the person being recognized, and is more readily accepted by the general public in comparison with other biometrics which are regarded as more intrusive. In addition, the acquisition of data for face recognition can be performed readily and cheaply, using widely available devices. However, although the interest in security uses of face recognition continues, it has become increasingly the case that research has shifted from the use of face recognition as a sole biometric. Instead, the operational paradigm is that of face recognition as an element of a multi-biometric system \cite{MuraIwamMakiYagi2013}.

Of particular interest to us is the use of infrared imaging as a modality complementary to the `conventional', visible spectrum-based face recognition. Our focus is motivated by several observations. Firstly, in principle an IR image of a face can be acquired whenever a conventional image of a face can. This may not be the case with other biometrics (gait, height etc). Secondly, while certainly neither as cheap nor as widely available as conventional cameras, in recent years IR imagers have become far more viable for pervasive use. It is interesting to note the self-enforcing nature of this phenomenon: the initial technology advancement-driven drop in price has increased the use of IR cameras thereby making their production more profitable and in turn lowering their cost even further.

In a distal sense, the key challenges in IR-based face recognition remain to be pose invariance, robustness to physiological conditions affecting facial IR emissions, and occlusions due to facial hair and accessories (most notably eyeglasses). In a more proximal sense, as argued in a recent review \cite{GhiaAranBendMald2013b}, the main challenge is to formulate a framework which is capable of dealing with all of the aforementioned factors affecting IR `appearance' in a unified manner. While a large number of IR-based face recognition algorithms have been described in the literature, without exception they all constrain their attention to a few, usually only a single, extrinsic factor (e.g.\ facial expression or pose). None of them can cope well with a concurrent variability in several extrinsic factors. Yet, this is the challenge encountered in practice.

In this paper our aim is to describe what we believe to be the first IR-based face recognition method which is able to deal with all of the major practical challenges. Specifically, our method explicitly addresses (i) the variability in the user's physiological state, (ii) pose changes, (iii) facial expressions, (iv) partial occlusion due to prescription glasses, and (v) quasi-occlusion due to facial hair.

\section{Unified treatment of extrinsic variability}\label{s:main}
In this section we detail different elements of our system. We start by describing the Dual Dimension Active Appearance Model Ensemble framework and then demonstrate how it can be extended to perform model selection which allows for the handling of partial occlusion due to prescription glasses, and quasi-occlusion due to facial hair.

\subsection{Dual Dimension AAM Ensemble (DDAE)}
At the coarsest level, the Dual Dimension Active Appearance Model Ensemble algorithm comprises three distinct components. These are: (i) a method for fitting an active appearance model (AAM) \cite{GrosMattBake2006} particularly designed for fast convergence and reliable fitting in the IR spectrum, (ii) a method for selecting the most appropriate AAM from a trained ensemble, and (iii) the underlying extraction of person-specific discriminative information. The ultimate functions of these elements within the system as a whole are respectively pose normalization within a limited yaw range, invariance across the full range of yaw, and invariance to physiological changes of the user.

\vspace{-10pt}
\subsubsection{AAM fitting}
\begin{SCfigure}
 \vspace{-20pt}
  \centering
  \includegraphics[width=0.4\textwidth]{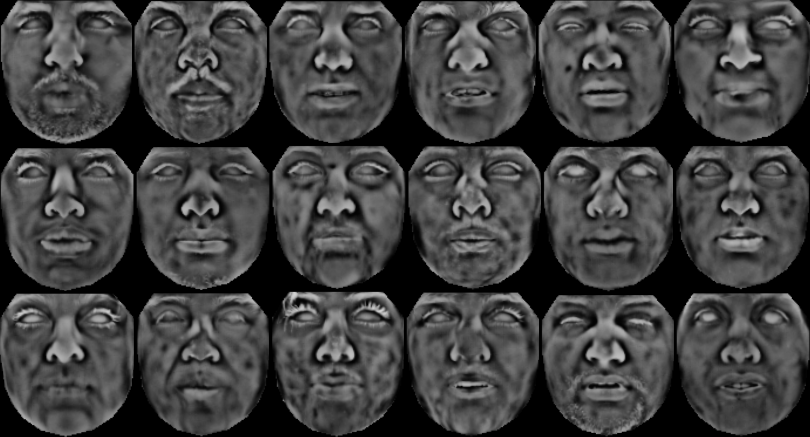}
  \caption{Examples of automatically pre-processed thermal images warped to the canonical geometric frame using the described AAM fitting method. }
  \label{f:aamExamp}
  \vspace{-10pt}
\end{SCfigure}

There are two crucial design aspects of the design and deployment of AAMs that need to be considered in order to ensure their robustness. These are: (i) the model initialization procedure, and (ii) the subsequent iterative refinement of model parameters. In the context of the problem considered in this paper, the former is relatively simple. Given that we are using thermal imaging background clutter is virtually non-existant applying simple thresholding to the input image allows the face to be localized and its spatial extent estimated. Reliable initialization of the AAM is then readily achieved by appropriately positioning its centroid and scale. A much greater challenge in the use of the AAM model for the normalization of pose and facial expression concerns the subsequent convergence -- the model is notoriously prone to convergence towards a local mininum, possibly far from the correct solution. This problem is even more pronounced when fitting is performed on face images acquired in the IR spectrum; unlike in the visible spectrum, in thermal IR human faces lack face-specific detail that is crucial in directing iterative optimization. This is the likely explanation for the absence of published work on the use of AAMs for faces in IR until the work of Ghiass \textit{et al.}~\cite{GhiaAranBendMald2013}. Their key idea was to perform fitting by learning and applying an AAM not on raw IR images themselves but rather on images automatically processed in a manner which emphasizes high-frequency detail. Specifically the detail-enhanced image $I_e$ is computed by anisotropically diffusing the input image $I$:
{\small\begin{align}
  \frac{\partial I}{\partial t} = \nabla.\left( c(\| \nabla I\|)~\nabla I\right) = \nabla c.\nabla I + c(\| \nabla I\|)~\Delta I,
\end{align}}
using a spatially varying and image gradient magnitude-dependent parameter $c(\| \nabla I\|) = \exp \left\{ -\|\nabla I\|/400 \right\}$, subtracting the result from the original image and applying histogram equalization $I_e = \text{histeq}(I - I_d)$. Warped examples are shown in Fig.~\ref{f:aamExamp}.

\subsubsection{Person and pose-specific model selection}
Applied to faces, the active appearance model comprises a triangular mesh with each triangle describing a small surface patch. To use this model for the normalization of pose, it is implicitly assumed that the surface patches are approximately planar. While this approximation typically does not cause fitting problems when pose variation is small this is not the case when fitting needs to be performed across a large range of poses (e.g.\ ranging from fully frontal to fully profile orientation relative to the camera's viewing direction). This is particularly pronounced when the applied AAM is generic (rather than person-specific) and needs to have sufficient power to describe the full scope of shape and appearance variability across faces. The DDAE method overcomes these problems by employing an ensemble of AAMs. Each AAM in an ensemble `specializes' to a particular region of IR face space. The space is effectively partitioned both by pose and the amount of appearance variability, making each AAM specific to a particular range of poses and individuals of relatively similar appearance (in the IR spectrum). In training, this is achieved by first dividing the training data corpus into pose-specific groups and then applying appearance clustering on IR faces within each group. A single AAM in an ensemble is trained using a single cluster. On the other hand, when the system is queried with a novel image containing a face in an arbitrary and unknown pose, the appropriate AAM from the ensemble needs to be selected automatically. This can be readily done by fitting each AAM from the ensemble and then selecting the best AAM as the one with the greatest maximal likelihood, that is, the likelihood achieved after convergence.

\vspace{-15pt}
\subsubsection{Discriminative representation}
\begin{figure}[thp]
  \centering
  \subfigure[Usr~1, seq~1]{~~\includegraphics[width=0.14\textwidth]{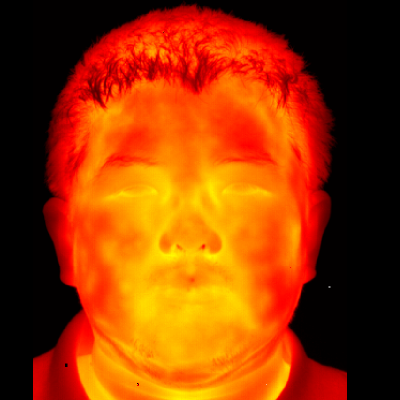}~~}
  \subfigure[Usr~1, seq~2]{~~\includegraphics[width=0.14\textwidth]{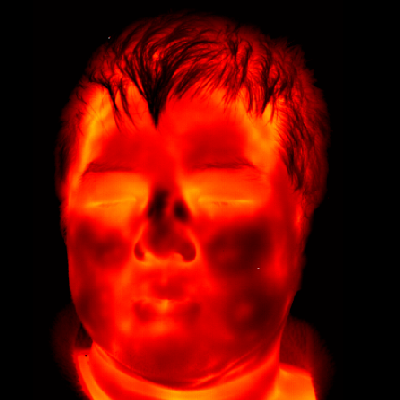}~~}~~~
  \subfigure[Usr~2, seq~1]{~~\includegraphics[width=0.14\textwidth]{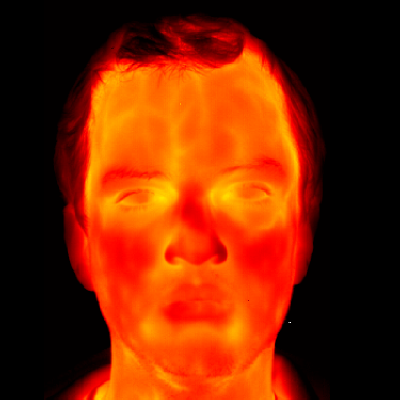}~~}
  \subfigure[Usr~2, seq~2]{~~\includegraphics[width=0.14\textwidth]{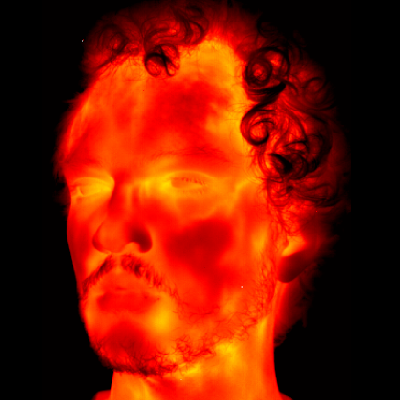}~~}
  \vspace{-5pt}
  \caption{Thermal IR appearance before (`seq 1') and after (`seq 2') mild exercise. }
  \label{f:phys}
  \vspace{-10pt}
\end{figure}

A major challenge to IR-based face recognition in practice emerges as a consequence of thermal IR appearance dependence on the physiological state of the user. Factors which affect the sympathetic or parasympathetic nervous system output (e.g.\ exercise or excitement) or peripheral blood flow (e.g.\ ambient temperature or drugs) can have a profound effect. This is illustrated in Fig.~\ref{f:phys}. With the notable exception of the work by Buddharaju \textit{et al.} \cite{BuddPavlTsiaBaza2007} there has been little research done on developing an IR-based person-specific representation invariant to these changes.

We adopt the use of a representation which is dependent on the distribution of superficial blood vessels. Unlike the vessel network based representation \cite{BuddPavlTsiaBaza2007} our representation is not binary i.e.\ a particular image pixel is not merely classified as a blood vessel or not. Rather, the extracted pseudo-probability map allows for the level of confidence regarding the saliency of a particular pixel to be encoded. Additionally, unlike the representation extracted by various blood perfusion methods \cite{SealNasiBhatBasu2011}, our representation is based on temperature gradients, rather than the absolute temperature. As such, it is less affected by physiological changes which influence the amount of blood flow, and is rather a function of the invariant distribution of underlying blood vessels.

Our representation is extracted using the so-called vesselness filter \cite{FranNiesVincVier1998}, originally developed for use on 3D MRI data for the extraction of tubular structures from images. Just as in 3D, in 2D this is achieved by considering the of the Hessian matrix $H(I,x,y,s)$ computed at the image locus $(x,y)$ and at the scale $s$:
{\small\begin{align}
  H(I,x,y,s)=\begin{bmatrix}
         L_x^2(x,y,\sigma)    & L_x L_y (x,y,\sigma)\\
         L_x L_y (x,y,\sigma) & L_y^2(x,y,\sigma) \\
       \end{bmatrix}
\end{align}}
where $L_x^2(x,y,s)$, $L_y^2(x,y,s)$, and $L_x L_y (x,y,\sigma)$ are the second derivatives of $L(x,y,s)$, resulting from the smoothing the original image $I(x,y)$ with a Gaussian kernel $L(x,y,s) = I(x,y) \ast G(s)$. If the two eigenvalues of $H(I,x,y,s)$ are $\lambda_1$ and $\lambda_2$ and if without loss of generality we take $|\lambda_1| \leq |\lambda_2|$, two statistics which characterize local appearance and which can be used to quantify the local vesselness of appearance are $\mathcal{R}_\mathcal{A} = |\lambda_1|/|\lambda_2|$ and $\mathcal{S} = \sqrt{\lambda_1^2 + \lambda_2^2}$. The former of these measures the degree of local 'blobiness' which should be low for tubular structures, while $\mathcal{S}$ rejects nearly uniform, uninformative image regions which are characterized by small eigenvalues of the Hessian. For a particular scale of image analysis $s$, the two measures, $\mathcal{R}_\mathcal{A}$ and $\mathcal{S}$, are unified into a single measure $\mathcal{V}(s) = (1-e^{-\frac{\mathcal{R}_\mathcal{A}}{2\beta^2}}) \times (1-e^{-\frac{\mathcal{S}}{2c^2}})$ for $\lambda_2 > 0$ and $\mathcal{V}(s) =0$ otherwise, where $\beta$ and $c$ are the parameters that control the sensitivity of the filter to $\mathcal{R}_\mathcal{A}$ and $\mathcal{S}$. The overall vesselness of a particular image locus can be computed as the maximal vesselness across scale $\mathcal{V}_0 = \max_{s_{min} \leq s \leq s_{max}} \mathcal{V}(s)$.

\subsection{Robustness to eye-wear and facial hair changes}
A significant challenge posed to face recognition algorithms, conventional and IR-based ones alike, is posed by occlusions \cite{GrosMattBake2006,Mart2002}. Of particular interest to us are specific commonly encountered occlusions -- these are occlusions due to prescription glasses \cite{HeoKongAbidAbid2004} (for practical purposes nearly entirely opaque to the IR frequencies in the short, medium and long wave sub-bands and facial hair. To prevent them having a dramatic effect on intra-personal matching scores, it is of paramount importance to detect and automatically exclude from comparison the corresponding affected regions of the face. The DDAE framework can be extended to achieve precisely this.

\begin{SCfigure}
  \centering
  \includegraphics[width=0.233\textwidth]{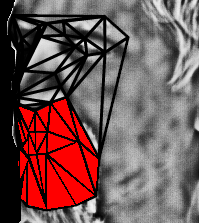}
  \caption{A geometrically truncated AAM. The truncated portion (red) is not used for fitting -- here it is shown using the average shape after the fitting of the remainder of the mesh.  }
  \label{f:mask}
\end{SCfigure}

In particular, we extend the existing AAM ensemble with additional models which are now occlusion-specific too. In the training stage this can be achieved by including AAMs which correspond to the existing ones but which are geometrically truncated. Note that this means that the new AAMs do not need to be re-trained -- it is sufficient to adopt the already learnt appearance and shape modes and truncate them directly. In particular, we created two truncated models -- one to account for the growth of facial hair (beard and moustache) and one to account for the presence of eye-wear. These two can also be combined to the produce the third truncated model for the handling of differential facial hair and eye-wear between two images hypothesized to belong to the same person. We created the two baseline truncated models manually; this is straightforward to do using high-level domain knowledge, as the nature of the specific occlusions in question constrains them to very specific parts of the face. An example of a fitted geometrically truncated AAM is shown in Fig.~\ref{f:mask}.\vspace{-12pt}

\vspace{-12pt}The application of the new ensemble in the classification of a novel image of a face, and in particular the process of model selection needed to achieve this, is somewhat more involved. In particular, the strategy whereby the highest likelihood model is selected is unsatisfactory as it favours smaller models (in our case the models which describe occluded faces). This is a well-known problem in the application of models which do not seek to describe jointly the entire image, i.e.\ both the foreground and the background, but the foreground only. Thus, we overcome this by not selecting the highest likelihood model but rather the model with the highest log-likelihood normalized by the model size \cite{AranCipo2006c}. Recall that the inverse compositional AAM fitting error is given by $e_{icaam} = \sum_{\text{All pixels } \mathbf{x}} \left[ I_e(\mathbf{W}(\mathbf{x};\mathbf{p})) - A_0(\mathbf{W}(\mathbf{x};\mathbf{p})) \right]^2$, where $\mathbf{x}$ are pixel loci, $A_i$ the retained appearance principal components and $\mathbf{W}(\mathbf{x};\mathbf{p})$ the location of the pixel warped using the shape parameters $\mathbf{p}$. Noting that by design of the model the error contributions of different pixels are de-correlated, in our case the highest normalized log-likelihood model is the one with the lowest mean pixel error $\bar{e_{icaam}}$.


\section{Evaluation and results}
In this section we describe the experiments we conducted with the aim of assessing the effectiveness of the proposed framework, and analyze the empirical results thus obtained. We start by summarizing the key features of our data set, follow that with a description of the evaluation methodology, and finally present and discuss our findings.

\vspace{-12pt}\subsubsection{Evaluation data}
In Sec.~\ref{s:intro} we argued that a major limitation of past research lies in the \textit{ad hoc} approach of addressing different extrinsic sources of IR appearance variability. Hand in hand with this goes the similar observation that at present there are few large, publicly available data sets that allow a systematic evaluation of IR-based face recognition algorithms and which contain a gradation of variability of all major extrinsic variables. We used the recently collected Laval University Thermal IR Face Motion Database \cite{GhiaAranBendMald2013a}. The database includes 200 people, aged 20 to 40, most of whom attended two data collection sessions separated by approximately 6 months. In each session a single thermal IR video sequence of the person was acquired using FLIR's Phoenix Indigo IR camera in the 2.5--5$\mu$m wavelength range. The duration of all video sequences in the database is 10~s and they were captured at 30~fps, thus resulting in 300 frames of $320 \times 240$ pixels per sequence. The imaged subjects were instructed to perform head motion that covers the yaw range from frontal ($0^\circ$) to approximately full profile ($\pm 90^\circ$) face orientation relative to the camera, without any special attention to the tempo of the motion or the time spent in each pose. The pose variability in the data is thus extreme. The subjects were also asked to display an arbitrary range of facial expressions. Lastly, a significant number of individuals were imaged in different physiological conditions in the two sessions. In the first session all individuals were imaged in a relatively inactive state, during the course of a sedentary working day and after a prolonged stay indoors, for the second session some users were asked to come straight from the exposure to cold ($<0^\circ$C) outdoors temperatures, alcohol intake, and/or exercise; see Fig.~\ref{f:phys}. In addition, individuals who wore prescription glasses in the first session were now asked to take them off. Several participants were also asked to allow for the growth of facial hair (beard, moustache) between the two sessions.

\vspace{-12pt}\subsubsection{Evaluation methodology}\label{ss:evalMethod}
We evaluated the proposed algorithm in a setting in which the algorithm is trained using only a single image in an arbitrary pose and facial expression. The querying of the algorithm using a novel face is also performed using a single image, possibly in a different pose and/or facial expression. Pose, facial expression changes and partial occlusion due to eye-wear or hair all present a major challenge to the current state of the art, and the consideration of all of the aforementioned in concurrence make our evaluation protocol extremely challenging (indeed, more so than any attempted by previous work), and, importantly, representative of the conditions which are of interest in a wide variety of practical applications. In an attempt to perform a comprehensive comparative evaluation we contacted a number of authors of previously proposed approaches. However none of them was able or willing to provide us with the source code or a working executable of their methods. Thus herein we constrain ourselves to the comparison of the proposed method with the DDAE algorithm which was compared with a thermal minutia points \cite{BuddPavlTsia2006} and vascular networks \cite{BuddPavl2009} methods in \cite{GhiaAranBendMald2013a}.

\subsection{Results}
The key results evaluation are summarized in Table~\ref{t:recognition} and Fig.~4. These show respectively the recognition rates achieved by our system in different experiments we conducted, and the receiver-operator characteristic curves corresponding to the recognition experiments in the presence of occlusion (in all subjects) due to facial hair or prescription glasses.

\begin{figure}[thp]
  \centering
  \vspace{-15pt}

  \begin{tabular}{ccccc}
    \rotatebox{90}{~~~Facial hair}~~~~&
    \includegraphics[width=0.19\textwidth]{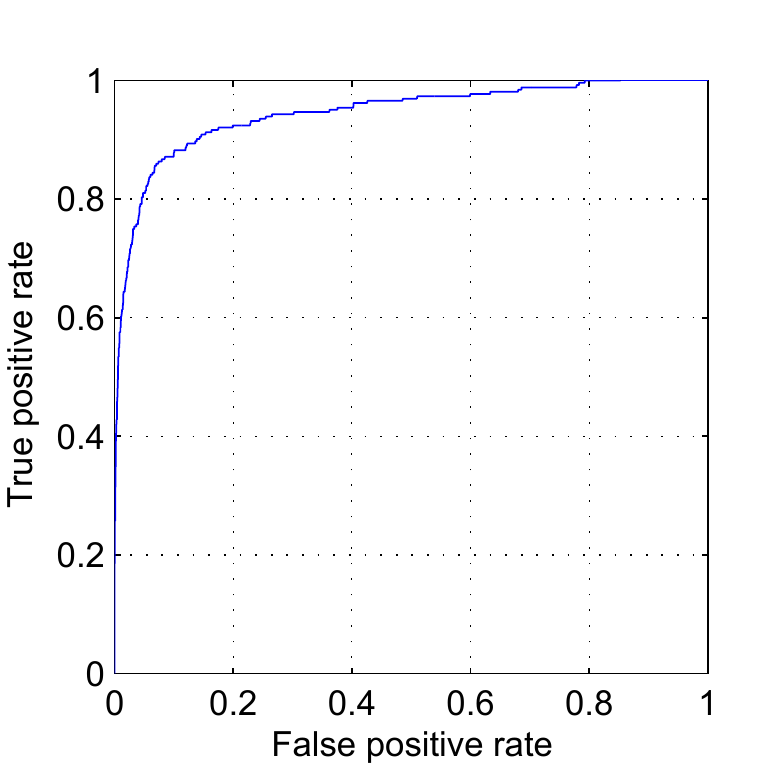} & \includegraphics[width=0.19\textwidth]{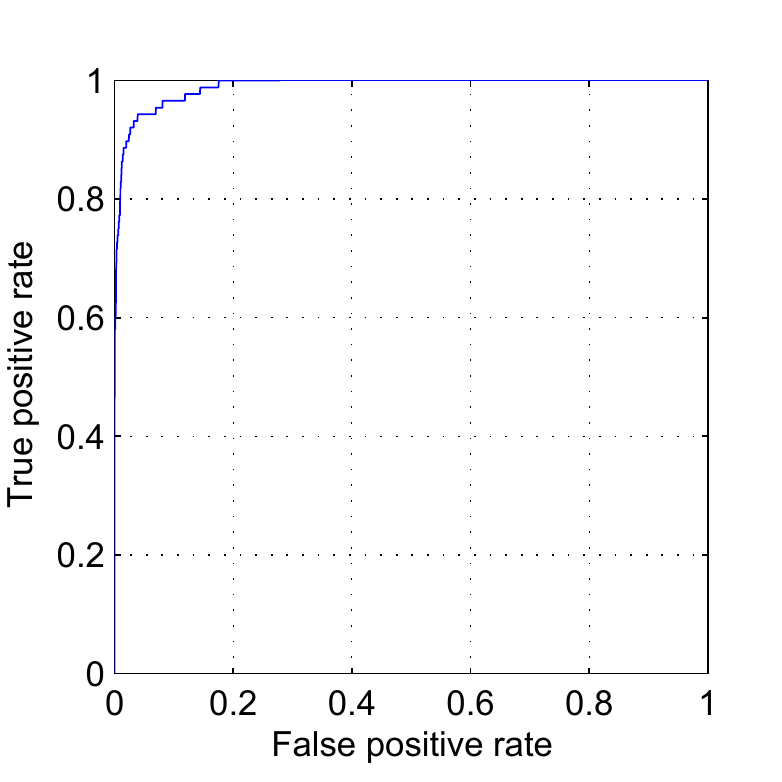} & \includegraphics[width=0.19\textwidth]{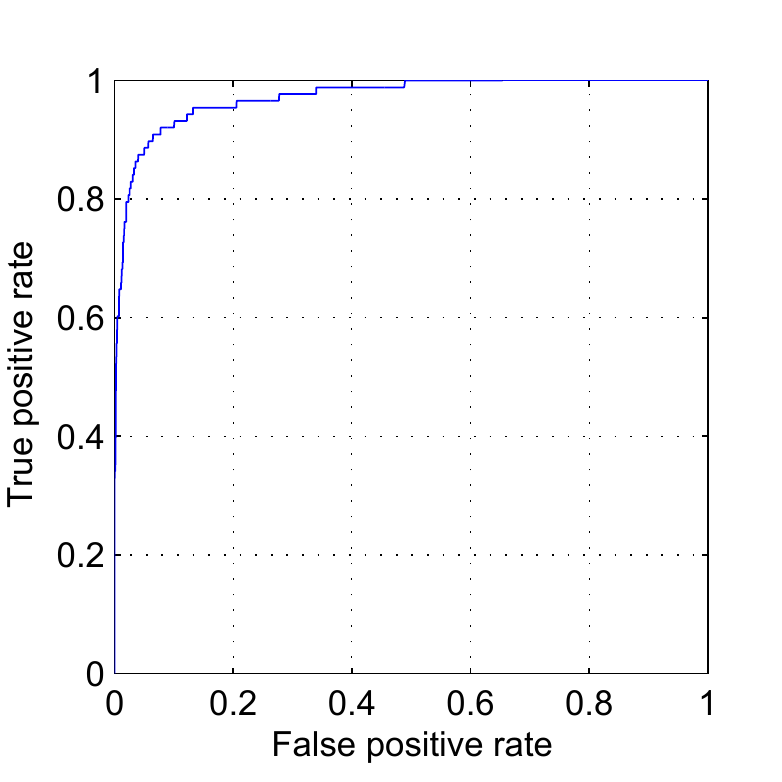} & \includegraphics[width=0.19\textwidth]{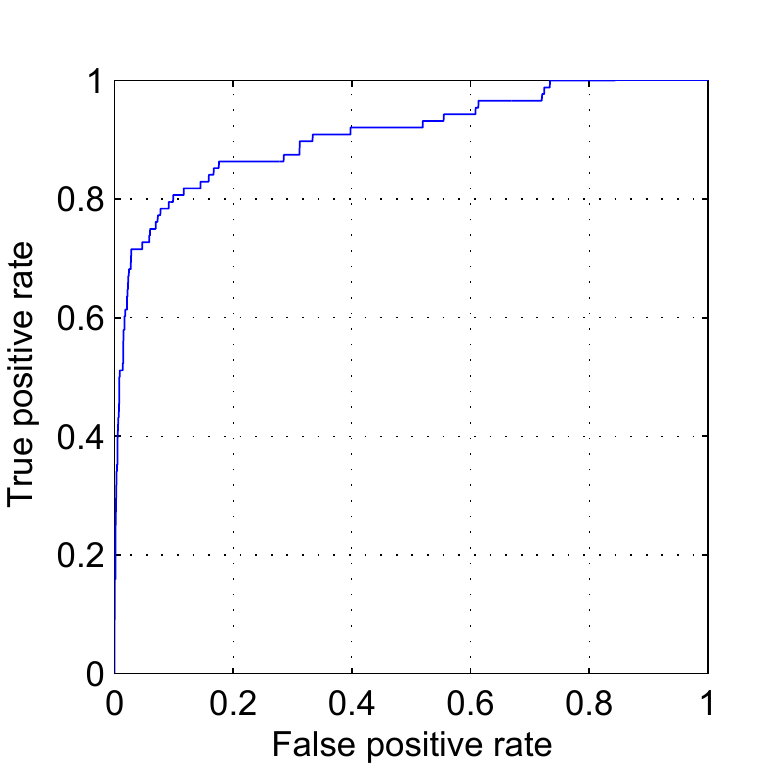}\\
    & \scriptsize Overall & \scriptsize $\Delta$ 0--30$^\circ$ & \scriptsize $\Delta$ 30--60$^\circ$ & \scriptsize $\Delta$ 60--90$^\circ$
  \end{tabular}
    \begin{tabular}{ccccc}
    \rotatebox{90}{~~~~~~Eye-wear}~~~~&
    \includegraphics[width=0.19\textwidth]{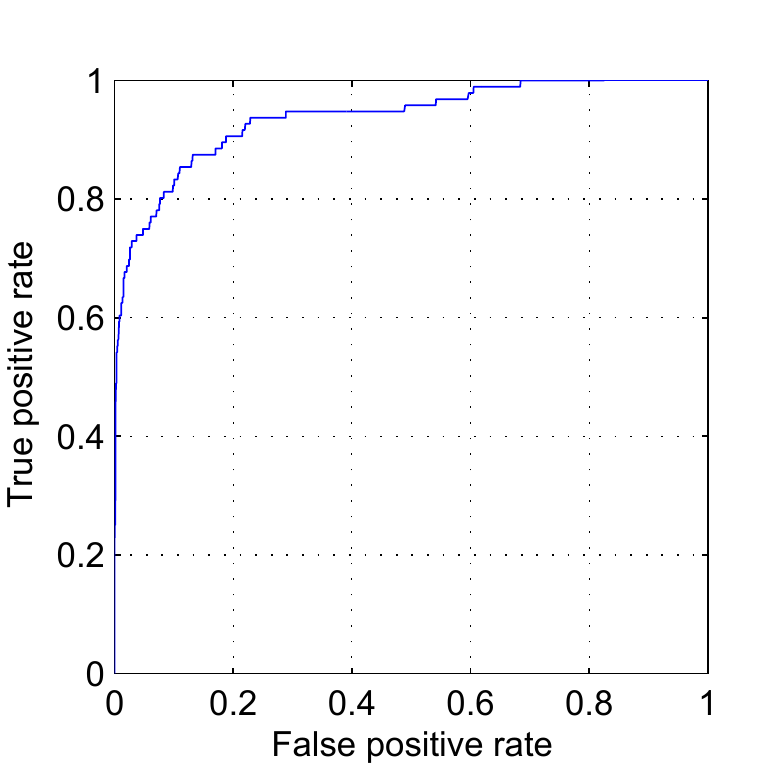} & \includegraphics[width=0.19\textwidth]{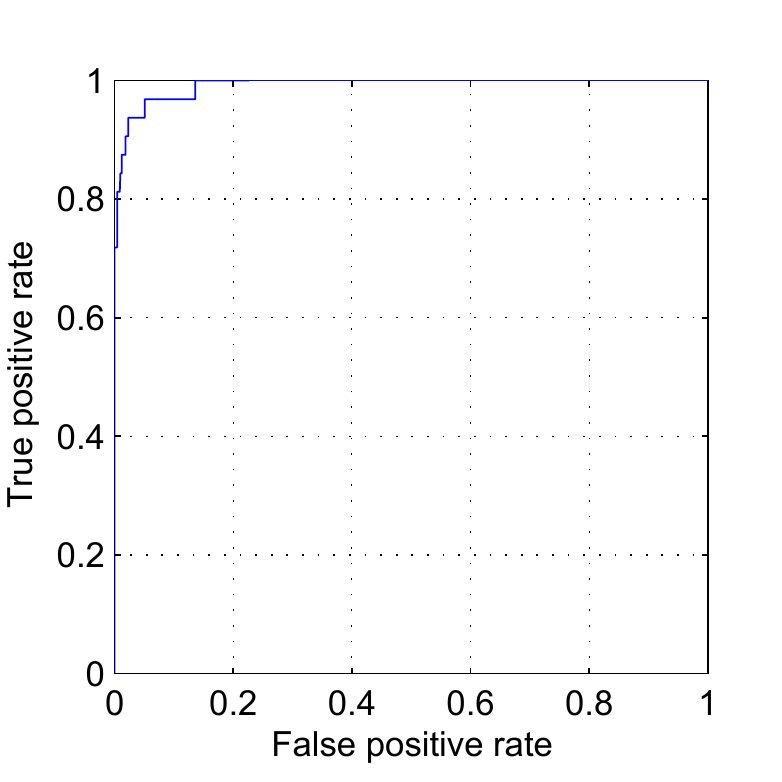} & \includegraphics[width=0.19\textwidth]{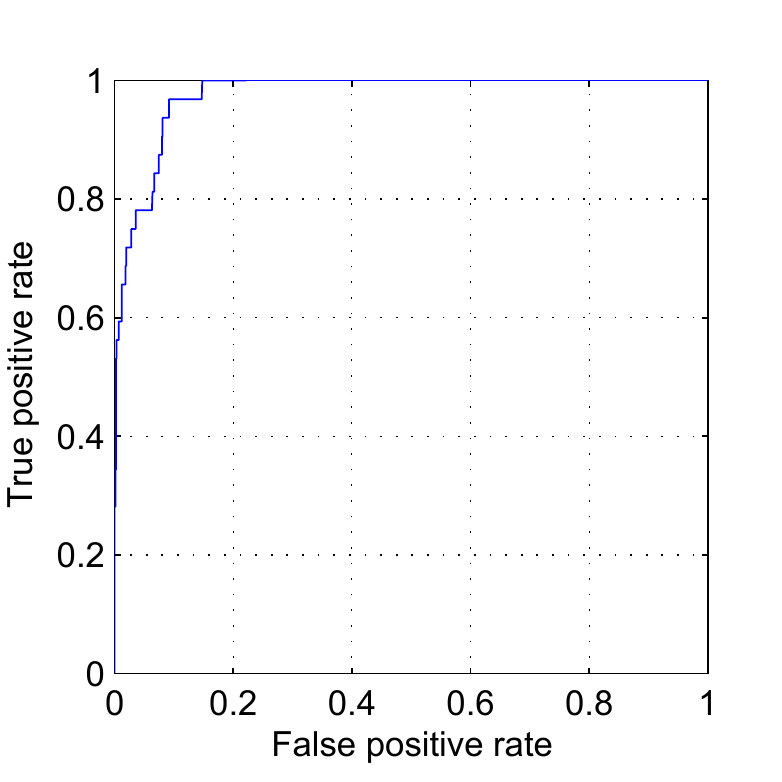} & \includegraphics[width=0.19\textwidth]{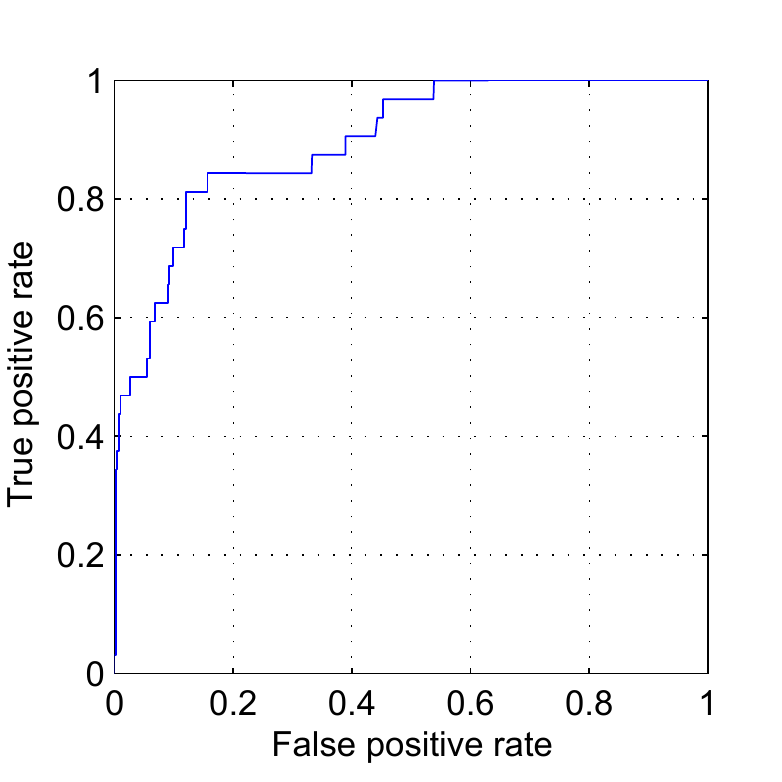}\\
    & \scriptsize Overall & \scriptsize $\Delta$ 0--30$^\circ$ & \scriptsize $\Delta$ 30--60$^\circ$ & \scriptsize $\Delta$ 60--90$^\circ$
  \end{tabular}
  \vspace{-5pt}
  \caption{Performance in the presence of occlusion across different extents of pose changes. }
  \label{f:roc}
  \vspace{-30pt}
\end{figure}

\begin{SCtable}
  \centering
  \small
  \vspace{-10pt}
  \caption{Average recognition rate. In experiments with partial occlusion the occlusion was differential (e.g.\ if a training image was acquired with eye-wear on, the test image was acquired with it off, and \textit{vice versa}.  }
  \renewcommand{\arraystretch}{1.2}
  \begin{tabular}{l|ccc}
    \Hline
                     & Unoccluded & Facial hair & Eye-wear\\
    \hline
    Rank 1 & 100\%        & 87\%  & 74\% \\
    Rank 2 & 100\%        & 94\%  & 84\% \\
    Rank 3 & 100\%        & 95\%  & 92\% \\
    \hline
    \Hline
  \end{tabular}
  \label{t:recognition}
  \vspace{-20pt}
\end{SCtable}

\vspace{-10pt}To start with, consider the results in Table~\ref{t:recognition}. It is interesting to notice that performance deterioration is greater when occlusion is caused by eye-wear, rather than facial hair growth. This may be particularly surprising considering that in our data the area occluded by facial hair was larger in extent. One possible explanation of this finding may be rooted in different discriminative abilities of different facial components. Further work is needed to ascertain this; although the eye region appears to be highly informative in the visible spectrum \cite{CampFeriCesa2000} this may not be the case in the IR spectrum as suggested by evidence from some previous work \cite{AranHammCipo2010}. However, there are alternative possibilities which may explain or contribute to the explanation of the observed performance differential. For example, the choice to grow a beard (say) is not arbitrary but rather a conscious decision made with aesthetic considerations in mind. It is possible that individuals who choose to grow facial hair have more characteristic faces.  It is also possible that the explanation is of a more practical nature -- perhaps the accuracy of AAM fitting is more affected by the absence of the information around the eyes, rather than those areas of the face typically covered by facial hair. We could not examine this quantitatively as it was prohibitively laborious to obtain the ground truth AAM parameters for the entire database. More research is certainly needed to establish the contribution of each of the aforementioned factors.

As Table~\ref{t:recognition} shows, both types of occlusions, those due to eye-wear and those due to facial hair, have a significant effect on recognition accuracy. However, what is interesting to observe is that already at rank-3 the correct recognition rate in all cases is at least 92\%. This exceeds the performance of the vascular networks based method which used thermal minutia points \cite{BuddPavlTsia2006} and is competitive with the iteratively registered networks approach \cite{BuddPavl2009}, even though the aforementioned algorithms employ several images per person for training and do not consider occlusions.

\section{Summary and conclusions}
We described what we believe to be the first attempt at addressing all major challenges in practical IR-based face recognition in a unified manner. In particular, our system explicitly handles changes in a person's pose, mild facial expression, physiological changes, partial occlusion due to eye-wear, and quasi-occlusion due to facial hair. Our future work will focus on the extension of the described framework to recognition from video and the utilization of partial information available in the regions of the face covered by facial hair.

\tiny
\bibliographystyle{unsrt}
\bibliography{./my_bibliography}

\begin{thebibliography}{10}

\bibitem{Shan2010}
C.~Shan.
\newblock {\em Video Search and Mining}, chapter Face Recognition and Retrieval
  in Video.
\newblock 2010.

\bibitem{ChelWilsSiro1995}
R.~Chellappa, C.~L. Wilson, and S.~Sirohey.
\newblock Human and machine recognition of faces: {A} survey.
\newblock {\em Proceedings of the IEEE}, 1995.

\bibitem{MuraIwamMakiYagi2013}
D.~Muramatsu, H.~Iwama, Y.~Makihara, and Y.~Yagi.
\newblock Multi-view multi-modal person authentication from a single walking
  image sequence.
\newblock {\em ICB}, 2013.

\bibitem{GhiaAranBendMald2013b}
R.~S. Ghiass, O.~Arandjelovi{\'c}, A.~Bendada, and X.~Maldague.
\newblock Infrared face recognition: a literature review.
\newblock {\em IJCNN}, 2013.

\bibitem{GrosMattBake2006}
R.~Gross, I.~Matthews, and S.~Baker.
\newblock Active appearance models with occlusion.
\newblock {\em IVC}, 2006.

\bibitem{GhiaAranBendMald2013}
R.~S. Ghiass, O.~Arandjelovi{\'c}, A.~Bendada, and X.~Maldague.
\newblock Vesselness features and the inverse compositional {AAM} for robust
  face recognition using thermal {IR}.
\newblock {\em AAAI}, 2013.

\bibitem{BuddPavlTsiaBaza2007}
P.~Buddharaju, I.~T. Pavlidis, P.~Tsiamyrtzis, and M.~Bazakos.
\newblock Physiology-based face recognition in the thermal infrared spectrum.
\newblock {\em PAMI}, 2007.

\bibitem{SealNasiBhatBasu2011}
A.~Seal, M.~Nasipuri, D.~Bhattacharjee, and D.K. Basu.
\newblock Minutiae based thermal face recognition using blood perfusion data.
\newblock {\em ICIIP}, 2011.

\bibitem{FranNiesVincVier1998}
A.~F. Frangi, W.~J. Niessen, K.~L. Vincken, and M.~A. Viergever.
\newblock Multiscale vessel enhancement filtering.
\newblock {\em MICCAI}, 1998.

\bibitem{Mart2002}
A.~M. Martinez.
\newblock Recognizing imprecisely localized, partially occluded and expression
  variant faces from a single sample per class.
\newblock {\em PAMI}, 2002.

\bibitem{HeoKongAbidAbid2004}
J.~Heo, S.~G. Kong, B.~R. Abidi, and M.~A. Abidi.
\newblock Fusion of visual and thermal signatures with eyeglass removal for
  robust face recognition.
\newblock {\em CVPRW}, 2004.

\bibitem{AranCipo2006c}
O.~Arandjelovi{\'c} and R.~Cipolla.
\newblock Automatic cast listing in feature-length films with anisotropic
  manifold space.
\newblock {\em CVPR}, 2006.

\bibitem{GhiaAranBendMald2013a}
R.~S. Ghiass, O.~Arandjelovi{\'c}, A.~Bendada, and X.~Maldague.
\newblock Illumination-invariant face recognition from a single image across
  extreme pose using a dual dimension {AAM} ensemble in the thermal infrared
  spectrum.
\newblock {\em IJCNN}, 2013.

\bibitem{BuddPavlTsia2006}
P.~Buddharaju, I.~Pavlidis, and P.~Tsiamyrtzis.
\newblock Pose-invariant physiological face recognition in the thermal infrared
  spectrum.
\newblock {\em CVPRW}, 2006.

\bibitem{BuddPavl2009}
P.~Buddharaju and I.~Pavlidis.
\newblock Physiological face recognition is coming of age.
\newblock {\em CVPR}, 2009.

\bibitem{CampFeriCesa2000}
T.~E. de~Campos, R.~S. Feris, and R.~M. Cesar~Junior.
\newblock Eigenfaces versus eigeneyes: First steps toward performance
  assessment of representations for face recognition.
\newblock {\em MICAI}, 2000.

\bibitem{AranHammCipo2010}
O.~Arandjelovi{\'c}, R.~I. Hammoud, and R.~Cipolla.
\newblock Thermal and reflectance based personal identification methodology in
  challenging variable illuminations.
\newblock {\em PR}, 2010.

\end{thebibliography}
\end{document}